\documentclass[letterpaper]{article} 
\usepackage[preprint]{ChronoScribe}
\usepackage[hyphens]{url}  
\usepackage{graphicx} 
\usepackage{natbib}  
\usepackage{caption} 
\usepackage{algorithm}
\usepackage{algorithmic}
\usepackage{booktabs}
\usepackage{amsmath}
\usepackage{amssymb}
\usepackage{tikz}
\usetikzlibrary{arrows.meta,positioning}
\title{ChronoScribe: Semantic--Numerical Specialization for\\Instruction-Driven Time Series Forecasting}
\author{
    Yuchen Jiao,
    Miao Zhang,
    Yuxuan Liu,
    Chengxi Li
}
\affiliations{Beijing University of Posts and Telecommunications\\
\url{https://github.com/Thekigs/ChronoScribe}}

\begin{document}

\maketitle

\begin{abstract}
Recent advances in instruction-driven time series forecasting highlight a fundamental tension: systems must simultaneously master natural-language instruction following and precise numerical prediction. Current approaches often force a compromise. Large language models (LLMs) can interpret unstructured requests but may be inaccurate when asked to produce forecast values directly, whereas time series foundation models (TSFMs) provide numerical forecasts but typically require structured inputs specifying the target series, context window, and horizon. We present \textbf{ChronoScribe}, which applies semantic--numerical specialization to this workflow. A LoRA-adapted LLM proposes an execution path and context length; deterministic rules resolve the executable configuration. On the specialist path, a frozen TSFM produces quantile trajectories for the task-specified horizon. Deterministic trajectory summaries condition the response, while the LLM neither generates nor revises predicted values. We evaluate a Qwen3.5-9B--Chronos-2 implementation across 23 evaluation dimensions and 30 public datasets against eleven proprietary direct LLMs and three open TSFMs. ChronoScribe has the lowest error on sMAPE, MASE, WAPE, RMSE, and MAE among all evaluated systems, including fixed-context controls. Across these metrics, even the lowest direct-LLM error remains $2.0$--$2.7\times$ the corresponding ChronoScribe error. At the system level, ChronoScribe achieves a Task-Level Score (TLS) of $65.5$, $10.6$ points above the strongest direct-LLM baseline. The results support ChronoScribe's division of labor, in which the LLM handles interpretation, configuration, and communication while the TSFM handles numerical forecasting.
\end{abstract}

\section{Introduction}

Time series forecasting is commonly formulated as the task of estimating future values from an observed history over a specified horizon. In practice, users may use natural language to request a forecast together with a trend summary or uncertainty statement. A request such as ``forecast demand for the coming week and summarize the trend and uncertainty'' requires a system to interpret the request, configure and run a forecaster, and report the result. Large language models (LLMs) and time series foundation models (TSFMs) can support this workflow, but they are suited to different parts of it.

TSFMs support zero-shot and probabilistic forecasting through large-scale pretraining~\cite{ansari2025chronos2}, but generally expect a structured interface that specifies the target series, horizon, and context window. LLMs can interpret open-ended instructions and produce natural-language responses, yet their training objectives do not directly optimize scale fidelity or temporal forecast accuracy~\cite{gruver2023llmtime,jin2024timellm}. Using a language model for both request interpretation and trajectory generation combines tasks with different modeling requirements. The resulting design problem is to let an LLM configure a numerical specialist without directly generating or revising the forecast trajectory.

Some prior approaches use language models to generate or represent time series; others use textual context to steer or revise forecasts~\cite{gruver2023llmtime,jin2024timellm,wang2025chronosteer,zhao2025timeseriesscientist,liao2026lastmile}. We focus on the interface between the two model classes. Such an interface must translate an open-ended request into a valid specialist call and communicate the result without becoming a second source of predicted values. Numerical provenance therefore depends on keeping the directive, specialist output, and response distinct.

\begin{figure}[t]
  \centering
  \includegraphics[width=\columnwidth]{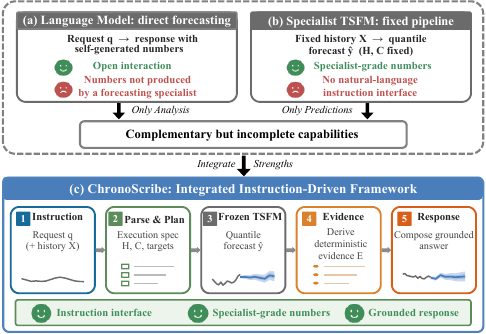}
  \caption{Direct LLMs, fixed TSFMs, and ChronoScribe.}
  \label{fig:teaser}
\end{figure}

Separating these stages makes numerical provenance easier to trace and failures easier to diagnose. A direct LLM must preserve scale, units, and long-range temporal structure while composing a fluent answer. A fixed-context specialist avoids direct language-model forecasting but cannot adapt its history length to individual requests. \mbox{Figure~\ref{fig:teaser}} contrasts direct LLMs, fixed TSFMs, and ChronoScribe. Allowing the language model to generate or revise forecast values would obscure their source. A typed interface therefore exposes the proposal, executable configuration, specialist output, and final response as separate artifacts.

We call this division of responsibility \emph{semantic--numerical specialization}. The LLM handles request interpretation, specialist configuration, and communication, whereas the TSFM alone generates forecast values. ChronoScribe implements this principle with a LoRA-adapted language model that emits a typed directive. The directive proposes a direct language path or a specialist path and, in the latter case, a context length. Deterministic rules parse and normalize the proposal or apply a fallback; raw language-model output is not executed directly. The task, rather than the directive, supplies the forecast horizon. A frozen TSFM then produces quantile trajectories from the selected history. Deterministic functions derive numerical summaries, a direction label, and uncertainty intervals from these trajectories. The adapted language model uses this evidence to compose the response without generating or revising the trajectory.

Our evaluation spans 23 dimensions in forecasting, temporal reasoning, instruction following, uncertainty, and efficiency, using 30 public datasets. We compare ChronoScribe with eleven proprietary direct LLMs, three open TSFMs, fixed-context controls, and component ablations. ChronoScribe has the lowest sMAPE, MASE, WAPE, RMSE, and MAE among all evaluated systems, including fixed-context controls. Across these metrics, the lowest direct-LLM error remains $2.0$--$2.7\times$ the corresponding ChronoScribe error. Its predefined Task-Level Score (TLS), a weighted aggregate of the five capability groups, is $65.5$, $10.6$ points above the strongest direct-LLM baseline. The controller returns a parseable directive with the protocol-defined route for $95.9\%$ of requests. In a rule-based audit of 100 specialist-path responses, $87.4\%$ of 230 claims agree with deterministic forecast evidence, and 89 responses pass every applicable check.

This work contributes (i) ChronoScribe, a framework for semantic--numerical specialization that assigns request interpretation and evidence-based communication to an adapted LLM while reserving trajectory generation for a frozen TSFM; (ii) typed execution and forecast-evidence interfaces that allow raw directives, executable configurations, specialist forecasts, and evidence-conditioned responses to be inspected separately; and (iii) a 23-dimension evaluation across 30 public datasets against eleven proprietary direct LLMs, three open TSFMs, fixed-context controls, and component ablations, together with audits of directive reliability and forecast-claim grounding.

\section{Related Work}

Transformer forecasters such as Informer, PatchTST, iTransformer, and TimesNet use sparse attention, patching, inverted tokenization, or multi-scale representations for long-horizon prediction~\cite{zhou2021informer,nie2023patchtst,liu2024itransformer,wu2023timesnet}. Time series foundation models, including Chronos, Chronos-2, TimesFM, Moirai, Timer, and MOMENT, are pretrained on heterogeneous series for zero-shot or transfer forecasting~\cite{ansari2024chronos,ansari2025chronos2,das2024timesfm,woo2024moirai,liu2024timer,goswami2024moment}. At inference, these models operate on a selected series with a specified context window and horizon rather than on an open-ended request.

Several approaches incorporate pretrained language models into numerical forecasting. LLMTime serializes observations as text-like tokens and generates continuations directly. GPT4TS, TEMPO, UniTime, Time-LLM, TimeCMA, and CALF use fine-tuning, prompting, reprogramming, or cross-modal alignment to adapt language models or their representations~\cite{gruver2023llmtime,zhou2023gpt4ts,cao2024tempo,liu2024unitime,jin2024timellm,liu2025timecma,liu2025calf}. M3Time, T3Time, and ExoTimer incorporate multimodal, multiscale, or exogenous information~\cite{jia2026m3time,chowdhury2026t3time,wu2026exotimer}. In these systems, language-model generation or representations contribute to the numerical forecast. ChronoScribe assigns request interpretation, specialist configuration, and response composition to the LLM, while the frozen TSFM alone produces the trajectory.

Tool use offers a precedent for separating language interpretation from external execution~\cite{schick2023toolformer,yao2023react}. ChronoSteer converts textual events into revision instructions that steer a TSFM~\cite{wang2025chronosteer}. TimeSeriesScientist assigns data diagnosis, model selection, forecasting, and reporting to specialized agents~\cite{zhao2025timeseriesscientist}. Last-mile forecasting agents retrieve contextual evidence and apply explicit revision actions to a forecasting backbone~\cite{liao2026lastmile}. In ChronoScribe, the LLM proposes whether and how to invoke a frozen specialist and communicates deterministic summaries of its output; it does not revise the specialist trajectory.

Established forecasting evaluations focus on numerical outputs, using point-error metrics and proper scoring rules~\cite{makridakis2020m4,hyndman2006mase,gneiting2007strictly}. GIFT-Eval broadens zero-shot TSFM evaluation across datasets, domains, frequencies, and forecast lengths~\cite{aksu2024gifteval}. These evaluations do not test request interpretation or whether a language response is supported by forecast evidence. Our protocol retains the raw forecasting metrics and adds request interpretation, instruction compliance, uncertainty communication, and execution cost.

\section{Problem Formulation}

\subsection{Instruction-driven Forecasting Setting}

Let $q$ denote a natural-language request and $X_{1:T}=(\mathbf{x}_1,\ldots,\mathbf{x}_T)$ the available history, where $\mathbf{x}_t\in\mathbb{R}^{d}$; setting $d=1$ gives the univariate case. Each task also supplies $\tau=(H_{\tau},j^{\star})$, comprising the forecast horizon $H_{\tau}$ and target index $j^{\star}$, with $j^{\star}=1$ for a univariate series. Given $(q,X_{1:T},\tau)$, the system decides whether to invoke a specialist and, if so, how much history to pass to it. The evaluation therefore holds the horizon fixed and measures routing and context configuration rather than autonomous horizon inference. A request may additionally specify trend, uncertainty, or response-format requirements.

\subsection{Forecast Specification and Evidence Construction}

We separate the language model's raw directive from the executable configuration. Define the target history as $X^{(j^{\star})}_{1:T}=(x_{1,j^{\star}},\ldots,x_{T,j^{\star}})$. Its deterministic representation $\rho(X^{(j^{\star})}_{1:T})$ contains the history length, summary statistics, a linear-trend estimate, a uniformly subsampled trajectory, and recent observations. Let $a_X\in\{0,1\}$ indicate whether this history satisfies the specialist's preprocessing requirements. The directive proposes a path and context length; it also repeats the supplied horizon for schema compliance. The latter field does not determine the executable horizon:
\begin{equation}
\begin{aligned}
\hat{s}
&=
\Psi_{\theta}\!\left(
q,\rho(X^{(j^{\star})}_{1:T}),\tau
\right),\\
(u,C)
&=
\Gamma\!\left(\hat{s};q,X^{(j^{\star})}_{1:T},\tau\right),
\qquad H=H_{\tau}.
\end{aligned}
\label{eq:controller}
\end{equation}
The specialist runs only when the resolved action selects it and the target history is admissible:
\begin{equation}
Y=
\begin{cases}
F_{\phi}\!\left(X^{(j^{\star})}_{T-C+1:T};H\right), & u=1\ \text{and}\ a_X=1,\\
\varnothing, & \text{otherwise},
\end{cases}
\label{eq:forecast_execution}
\end{equation}
When $Y\neq\varnothing$, the deterministic map $e=g(Y)$ formats the forecast into an evidence record, and the language model samples $r\sim p_{\theta}(\cdot\mid q,e)$. When $Y=\varnothing$, the language model instead samples $r\sim p_{\theta}(\cdot\mid q)$. Here, $\hat{s}$ is the JSON-formatted directive, and $(u,C)\in\mathcal{Z}$ is the resolved action. The variable $u\in\{0,1\}$ selects the path, while $C\in\mathbb{N}_{+}$ is clipped to the admissible range for the available history and forecaster. The deterministic operator $\Gamma$ parses and normalizes $\hat{s}$ or applies a fallback if parsing fails. The repeated horizon field is audited for compliance, but $H_{\tau}$ always determines execution. Only the frozen forecaster $F_{\phi}$ generates future values; the language model produces either an evidence-conditioned response or a direct response.

\subsection{Forecast-grounded Response Generation}

For each forecast-claim type $k\in\mathcal{K}=\{\mathrm{num},\mathrm{dir},\mathrm{unc}\}$, let $\mathrm{Claims}_{k}(r)$ denote the numerical, directional, or uncertainty claims expressed in the specialist-path response $r$, and let $\mathrm{Derivable}_{k}(e)$ denote the corresponding statements supported by the forecast evidence. We call a response \emph{forecast-grounded} when
\begin{equation}
\mathrm{Claims}_{k}(r)
\subseteq
\mathrm{Derivable}_{k}(e),
\qquad
\forall k\in\mathcal{K}.
\label{eq:grounding}
\end{equation}
Equation~\eqref{eq:grounding} is the traceability test used in our audit, not a constraint enforced at runtime. Grounding does not require the response to reproduce the complete trajectory; it requires every forecast-related claim to be supported by a deterministic transformation of the specialist output. The test measures agreement with specialist-derived evidence, not forecast accuracy or calibration. It is applied only to the specialist path because a direct language response has no forecast evidence.

\section{ChronoScribe Framework}

\subsection{Overview}

\begin{figure*}[t]
  \centering
  \includegraphics[width=0.92\textwidth]{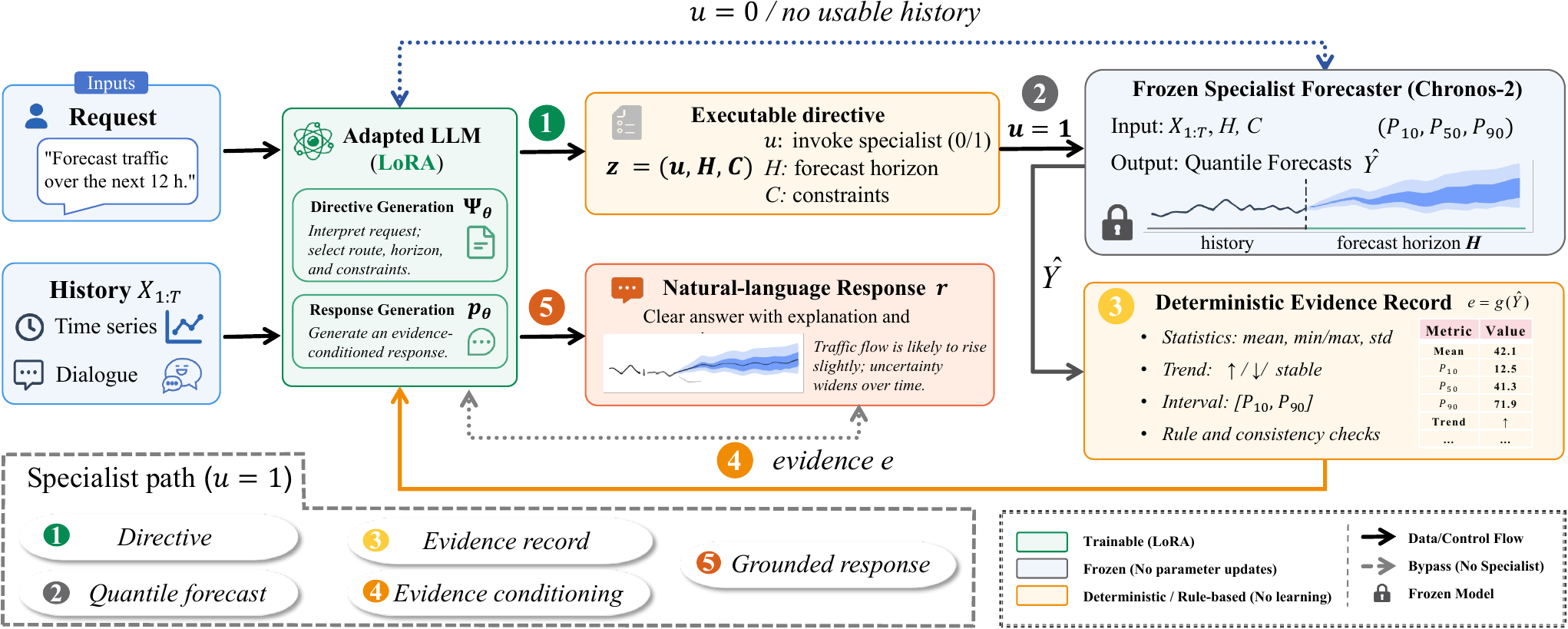}
  \caption{Typed-directive workflow with deterministic evidence.}
  \label{fig:framework}
\end{figure*}

ChronoScribe supports a direct language path and a specialist path. \mbox{Figure~\ref{fig:framework}} outlines this workflow. Every request first passes through the adapted language model, which emits a structured directive that deterministic rules resolve into a typed action. On the direct path, the model answers without invoking the forecaster. On the specialist path, a frozen forecaster receives the selected history window, and deterministic functions convert its quantile trajectories into an evidence record for the response. The same adapted language model emits the directive and composes the final response; the forecaster remains fixed.

\subsection{Routing and Forecast Specification}

The language component produces the raw directive $\hat{s}$ in Eq.~\eqref{eq:controller}. The parser $\mathcal{P}$ maps it to $\tilde{z}=\mathcal{P}(\hat{s})$, with $\tilde{z}=\bot$ denoting failure. If parsing succeeds, $\Gamma$ applies the normalizer $\mathcal{N}(\tilde{z};T,\tau)$, which casts fields to their required types and clips $C$ to the admissible range. If parsing fails, $\Gamma$ instead applies the deterministic fallback $\mathcal{B}(q,X^{(j^{\star})}_{1:T},\tau)$, which uses a predefined route and context policy. Both cases return the same typed action $(u,C)$; raw language-model output is never executed directly. A specialist action is executed only when the historical observations also pass the validity check $a_X$. Parsing failures are counted as controller failures even if the fallback allows the request to complete.

The resolved action controls only the route and the history suffix supplied to the specialist. It cannot change the externally specified horizon $H_\tau$ or the specialist parameters $\phi$; $F_{\phi}$ produces all forecast values.

\subsection{Specialist Forecasting, Evidence, and Response}

On the specialist path, ChronoScribe extracts the selected history window of the target variable, $X^{(C,j^{\star})}=X^{(j^{\star})}_{T-C+1:T}$. In our implementation, the frozen specialist receives this subsequence and returns quantile trajectories over $H$ future steps:
\begin{equation}
\begin{aligned}
Y
&=
F_{\phi}\!\left(X^{(C,j^{\star})};H\right)
\in
\mathbb{R}^{|\mathcal{A}|\times H},\\
Y_{\alpha,h}
&=
\hat{x}_{T+h,j^{\star}}^{(\alpha)},
\qquad
\alpha\in\mathcal{A},
\quad
h\in\{1,\ldots,H\}.
\end{aligned}
\label{eq:quantile_forecast}
\end{equation}
The symbol $\phi$ denotes the frozen specialist parameters, $\mathcal{A}$ is the set of quantile levels, and $Y_{\alpha,h}$ is the $\alpha$-quantile prediction for the target variable at step $h$. Our Chronos-2 instantiation uses $\mathcal{A}=\{0.1,0.5,0.9\}$, corresponding to P10, P50, and P90. The forecaster receives the task-specified univariate subsequence. The framework requires the specialist to accept the selected history and horizon and return at least a point forecast. Probabilistic output is needed only to construct uncertainty evidence.

The deterministic map $g$ converts the quantile trajectories into the fixed evidence record
\[
e=g(Y)
=
\bigl(
e_{\mathrm{num}},
e_{\mathrm{dir}},
e_{\mathrm{unc}},
e_{\mathrm{aux}}
\bigr).
\]
The component $e_{\mathrm{num}}$ stores the initial, final, mean, minimum, and maximum P50 values together with relative change. The component $e_{\mathrm{dir}}$ records a trend label derived using fixed thresholds, and $e_{\mathrm{unc}}$ stores interval information from P10 and P90. The optional $e_{\mathrm{aux}}$ holds task-specific summaries such as peak properties; $e_{\mathrm{aux}}=\varnothing$ when none is required. The grounding analysis covers numerical, directional, and uncertainty claims; auxiliary claims are outside its scope. For multivariate inputs, every summary refers to the task-specified target variable. Quantile-ordering checks, direction thresholds, and peak-detection rules are fixed before evaluation and documented in the supplementary protocol.

\begin{figure}[t]
  \centering
  \includegraphics[width=\columnwidth]{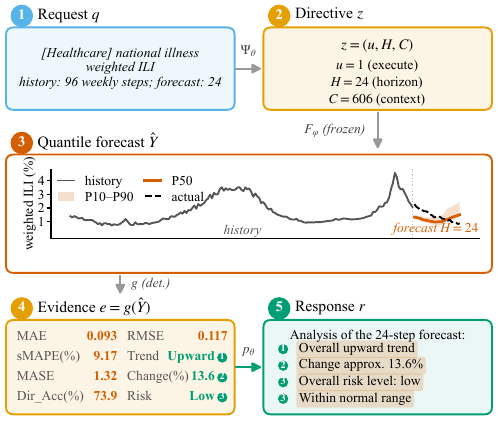}
  \caption{End-to-end specialist-path example.}
  \label{fig:case}
\end{figure}

The adapted language model generates a response conditioned on the original request and the evidence record. \mbox{Figure~\ref{fig:case}} gives a specialist-path example. Because the response remains free-form text, separating the language model from trajectory generation does not by itself guarantee Eq.~\eqref{eq:grounding}; the grounding audit checks each forecast-related claim against the evidence.

\subsection{Language Model Adaptation}

We train the language component on two interface tasks. A directive-generation record uses a request, history representation, and task constraints as input and a structured directive containing the route and context length as its target. A response-generation record uses a request and forecast evidence as input and an evidence-supported response as its target. Both tasks update a single LoRA adapter~\cite{hu2022lora}; the resulting model produces both directives and evidence-conditioned responses.

Let $c_i=(q_i,\rho_i,\tau_i)$ denote a directive input, where $\rho_i$ is the deterministic history representation defined above. Let $\mathcal{D}_{z}=\{(c_i,z_i)\}_{i=1}^{n_z}$ denote the directive-generation examples and $\mathcal{D}_{r}=\{\bigl((q_j,e_j),r_j\bigr)\}_{j=1}^{n_r}$ the evidence-to-response examples. We combine them as $\mathcal{D}=\mathcal{D}_{z}\cup\mathcal{D}_{r}$ and optimize only the LoRA parameters $\theta$ using the target-token-averaged autoregressive objective
\begin{equation}
\mathcal{L}_{\mathrm{SFT}}(\theta)
=
-\frac{1}{N}
\sum_{(a,b)\in\mathcal{D}}
\sum_{t=1}^{|b|}
\log
p_{\theta}
\left(
b_t\mid a,b_{<t}
\right),
\label{eq:sft_objective}
\end{equation}
where $(a,b)=(c_i,z_i)$ for directive-generation examples and $(a,b)=((q_j,e_j),r_j)$ for evidence-to-response examples. The normalization factor $N=\sum_{(a,b)\in\mathcal{D}}|b|$ is the total number of supervised target tokens. Prompt and padding tokens are masked from the loss. The adapter is trained only for directive and response generation; future numerical trajectories never appear as language-model targets.

The adaptation records use public time series datasets that are disjoint from the 30 evaluation datasets. No evaluation series, windows, or future targets are used during training. The supplementary material documents the source datasets, split rules, prompt templates, transformation procedures, and output schemas used to construct the records.

LoRA is applied to all attention and MLP projection layers with rank $64$, scaling parameter $128$, and dropout $0.05$. Training uses AdamW for two epochs with an effective batch size of $32$, a learning rate of $2\times10^{-4}$, weight decay $0.01$, and a $5\%$ linear warmup followed by cosine decay. The random seed is $42$. The base Qwen3.5-9B model and the Chronos-2 forecaster remain frozen throughout training.

\section{Experiments}

\subsection{Experimental Setup and Evaluation}

Our evaluation covers request interpretation, forecast execution, and response generation. The suite comprises 23 evaluation dimensions divided into five groups, with ten forecasting, six temporal-reasoning, five instruction-following, one uncertainty, and one efficiency dimension. The forecasting dimensions measure point accuracy, direction, peak behavior, cross-domain transfer, robustness, and horizon degradation. The remaining dimensions evaluate temporal judgments, instruction compliance, probabilistic communication, and inference cost.

The benchmark pool contains 8,120 source instances drawn from 30 public datasets in 14 families spanning energy, transportation, weather, finance, healthcare, retail, and monitoring. Primary comparisons use a fixed manifest of 460 records, with 20 records per dimension selected once with seed 2026. Each record contains a request $q$ and an observed history $X_{1:T}$; the future segment $X^{\star}_{T+1:T+H}$ is reserved for scoring. Every system receives the same request, history, and output requirements.

All benchmark prompts use Simplified Chinese; cross-lingual transfer is outside the present evaluation. Dataset provenance, preprocessing, templates, and scoring rules are provided in the supplementary material.

We report task-specific metrics before computing aggregate scores. The five primary point-error metrics are sMAPE, MASE, WAPE, RMSE, and MAE; directional and peak-oriented measures serve as additional forecasting diagnostics. Uncertainty is evaluated with interval coverage, calibration error, quantile-based criteria, and the continuous ranked probability score (CRPS). Temporal reasoning and instruction following are scored using task correctness, request interpretation, format validity, and required-field completeness. Efficiency is measured by latency, memory, throughput, and token consumption. Missing required outputs count as failures rather than being imputed. For the paired sMAPE comparison in the main results, we form a $95\%$ confidence interval from 1,000 dimension-level bootstrap replicates.

For a bounded higher-is-better metric $m\in[0,1]$, the score is $100m$. A lower-is-better metric is scored with the clipped linear map $100\max(0,1-m/b_m)$; an unbounded higher-is-better metric uses $100\min(1,\max(0,m/b_m))$. A reference bound $b_m$ is fixed for each metric before systems are compared. Boolean criteria receive $100$ for success and $0$ for failure. Each dimension score combines its applicable submetrics using fixed weights. The supplementary protocol lists every raw metric, reference bound, intra-dimension weight, direction of preference, and group assignment.

After mapping each evaluation dimension to $[0,100]$, we average the dimension scores within each group without additional weighting. Let $S_{\mathrm{F}}$, $S_{\mathrm{R}}$, $S_{\mathrm{I}}$, $S_{\mathrm{U}}$, and $S_{\mathrm{E}}$ denote the resulting forecasting, temporal-reasoning, instruction-following, uncertainty, and efficiency group scores. The Task-Level Score (TLS) is
\begin{equation}
\mathrm{TLS}
=
0.60S_{\mathrm{F}}
+
0.15S_{\mathrm{R}}
+
0.15S_{\mathrm{I}}
+
0.05S_{\mathrm{U}}
+
0.05S_{\mathrm{E}}.
\label{eq:tls}
\end{equation}
We fix the weights in Eq.~\eqref{eq:tls} before comparing systems. Numerical forecasting receives $60\%$ because it is the benchmark's central task. Temporal reasoning and instruction following each receive $15\%$; uncertainty communication and efficiency each receive $5\%$. TLS is the primary system-level metric, while raw metrics and group scores retain capability-level detail. A higher TLS does not indicate better performance in every group. In the sensitivity analysis, we vary the forecasting weight from $0.40$ to $0.80$ and allocate the remainder to reasoning, instruction following, uncertainty, and efficiency in the fixed ratio $3{:}3{:}1{:}1$. Setting $w_{\mathrm{F}}=0.60$ recovers the primary weights.

Our implementation pairs a LoRA-adapted Qwen3.5-9B language component~\cite{qwen35} with a frozen Chronos-2 forecaster. The comparisons include eleven proprietary direct LLMs, three open TSFMs (TimesFM, Moirai, and MOMENT), fixed-context controls, and component ablations. Each direct LLM receives the same request and history and must return the requested forecast or forecast-related response without access to Chronos-2 or the evidence module. API baselines use temperature $0$ where supported, a maximum output budget of 8,192 tokens, and a 300-second timeout.

Five fixed-context controls retain the language and specialist components but replace the proposed context length with a fixed global $C\in\{64,128,256,512,1024\}$. The \emph{w/o LoRA} variant keeps the two-path framework and substitutes the unadapted language model. The \emph{w/o Chronos-2} variant removes the specialist and forecasts directly with the language model. The \emph{C2-only} variant runs Chronos-2 with $C=512$ but omits language-guided routing, configuration, and response generation. These are comparisons between system variants, not estimates of additive component effects.

All systems are evaluated on the same records, with future targets reserved for scoring. The same parsers and scoring rules process every system's output. Model identifiers, prompts, inference settings, and access dates are listed in the supplementary material. Each system is evaluated three times on the fixed manifest. The bootstrap interval reported below resamples dimensions rather than repeated runs.

\subsection{Reproducibility}

Training and local inference use an NVIDIA A800. The artifact contains code, the fixed evaluation manifest, and reference configurations for adaptation and reported variants. The supplement specifies hardware/software details, data provenance and splits, prompts, transformations, output schemas, and evidence-verification rules.

\subsection{Main Results}

\begin{figure}[!t]
\centering
\includegraphics[width=\columnwidth]{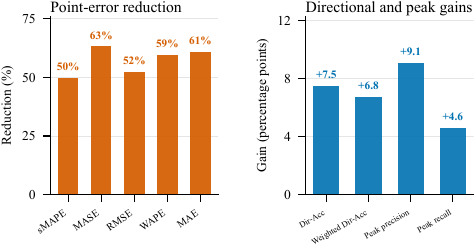}
\caption{Per-metric gains over the best direct LLM.}
\label{fig:forecast-gains}
\end{figure}

ChronoScribe achieves the lowest error on each of the five primary point-error metrics (sMAPE, MASE, WAPE, RMSE, and MAE) among all evaluated systems, including the fixed-context controls. \mbox{Table~\ref{tab:forecast-metrics}} reports the raw errors. Across these metrics, the lowest direct-LLM error remains $2.0$--$2.7\times$ the corresponding ChronoScribe error. This range represents relative reductions of approximately $50$--$63\%$. \mbox{Figure~\ref{fig:forecast-gains}} shows these gains. Relative to GPT-5.5~\cite{openai2026gpt55}, the best direct-LLM baseline on aggregate sMAPE, ChronoScribe lowers the metric from $50.85\%$ to $25.51\%$. The $95\%$ dimension-level bootstrap CI for this reduction is $[17.4,35.2]$ percentage points.

\begin{figure}[!t]
\centering
\includegraphics[width=\columnwidth]{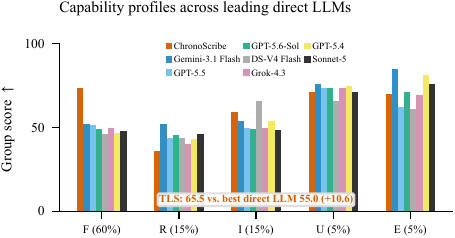}
\caption{Group scores: ChronoScribe vs. direct LLMs.}
\label{fig:capability-profiles}
\end{figure}

Using unrounded scores, ChronoScribe attains a TLS of $65.5$, $10.6$ points above Gemini-3.1-Flash~\cite{google2026gemini31flash} (TLS $55.0$). \mbox{Table~\ref{tab:system}} reports the system scores; \mbox{Figure~\ref{fig:capability-profiles}} profiles the five groups.

\begin{table}[!t]
\centering
\caption{System scores; column maxima are bold.}
\label{tab:system}
\footnotesize
\renewcommand{\arraystretch}{0.95}
\setlength{\tabcolsep}{1.4pt}
\begin{tabular*}{\columnwidth}{@{\extracolsep{\fill}}lrrrrrr@{}}
\toprule
Model & TLS & F & R & I & U & E \\
\midrule
\textbf{ChronoScribe} & \textbf{65.5} & \textbf{73.7} & 36.2 & 59.0 & 71.2 & 69.7 \\
\addlinespace[1pt]
\multicolumn{7}{l}{\emph{Direct LLMs}} \\
Gemini-3.1 Flash & 55.0 & 51.8 & \textbf{52.1} & 53.8 & 75.7 & 84.9 \\
GPT-5.5 & 51.7 & 51.5 & 43.5 & 49.8 & 73.6 & 62.3 \\
GPT-5.6-Sol & 50.9 & 49.1 & 45.5 & 49.3 & 73.5 & 71.3 \\
DeepSeek-V4 Flash & 50.3 & 46.0 & 43.5 & \textbf{65.8} & 65.7 & 61.1 \\
Grok-4.3 & 50.3 & 49.4 & 40.2 & 49.7 & 73.5 & 69.2 \\
GPT-5.4 & 50.3 & 46.5 & 43.3 & 53.7 & 74.9 & 81.2 \\
Claude Sonnet-5 & 50.0 & 47.6 & 46.1 & 48.2 & 70.9 & 75.8 \\
MiniMax-M3 & 50.0 & 51.8 & 40.2 & 44.7 & 62.5 & 61.6 \\
GLM-5.2 & 49.3 & 47.9 & 47.8 & 48.6 & 69.6 & 51.9 \\
DeepSeek-V4 Pro & 47.3 & 40.0 & 44.4 & 64.7 & 68.9 & 70.1 \\
Claude Opus-4.8 & 46.2 & 41.1 & 40.3 & 50.9 & 72.3 & 84.6 \\
\addlinespace[1pt]
\multicolumn{7}{l}{\emph{TSFMs}} \\
TimesFM & 49.1 & 44.3 & 26.8 & 63.9 & \textbf{82.9} & 95.1 \\
Moirai & 47.3 & 41.2 & 26.9 & 64.1 & 82.1 & 96.7 \\
MOMENT & 47.2 & 41.1 & 26.8 & 64.1 & 80.3 & \textbf{96.8} \\
\bottomrule
\end{tabular*}
\end{table}

\begin{table}[!t]
\centering
\caption{Point-forecast errors; column minima are bold.}
\label{tab:forecast-metrics}
\footnotesize
\renewcommand{\arraystretch}{0.95}
\setlength{\tabcolsep}{1.4pt}
\begin{tabular*}{\columnwidth}{@{\extracolsep{\fill}}lrrrrr@{}}
\toprule
Model & sMAPE & MASE & RMSE & WAPE & MAE \\
\midrule
\textbf{ChronoScribe} & \textbf{25.5} & \textbf{4.20} & \textbf{1.77} & \textbf{24.4} & \textbf{1.13} \\
\addlinespace[1pt]
\multicolumn{6}{l}{\emph{Direct LLMs}} \\
Gemini-3.1 Flash & 62.4 & 24.23 & 6.14 & 102.1 & 4.36 \\
GPT-5.5 & 50.9 & 14.81 & 3.70 & 60.3 & 2.87 \\
GPT-5.6-Sol & 57.3 & 14.35 & 3.93 & 68.0 & 3.51 \\
DeepSeek-V4 Flash & 62.7 & 13.80 & 4.46 & 70.3 & 3.62 \\
Grok-4.3 & 64.2 & 16.28 & 5.14 & 82.9 & 3.77 \\
GPT-5.4 & 63.4 & 15.68 & 4.13 & 74.9 & 3.48 \\
Claude Sonnet-5 & 65.0 & 15.01 & 4.53 & 72.6 & 3.80 \\
MiniMax-M3 & 57.5 & 13.20 & 4.62 & 60.6 & 3.52 \\
GLM-5.2 & 62.1 & 11.43 & 4.38 & 66.2 & 3.51 \\
DeepSeek-V4 Pro & 66.7 & 15.29 & 4.28 & 73.2 & 3.68 \\
Claude Opus-4.8 & 66.2 & 16.54 & 5.86 & 92.7 & 6.00 \\
\addlinespace[1pt]
\multicolumn{6}{l}{\emph{TSFMs}} \\
TimesFM & 69.7 & 16.12 & 8.54 & 105.0 & 7.59 \\
Moirai & 78.0 & 16.78 & 9.16 & 113.1 & 8.12 \\
MOMENT & 77.2 & 16.70 & 8.88 & 116.0 & 7.85 \\
\bottomrule
\end{tabular*}
\end{table}

The raw error ratios and the TLS margin summarize performance on different scales. Point-error ratios are computed directly from the original metrics. TLS instead maps heterogeneous metrics to $[0,100]$ using fixed reference bounds, averages dimension scores within each group, and applies the predefined $60/15/15/5/5$ weights. This process converts multiplicative forecasting-error gaps into bounded additive score differences, which are further attenuated by the $0.60$ forecasting weight. A $10.6$-point TLS lead is therefore compatible with the $2.0$--$2.7\times$ raw error ratios: it summarizes the complete workflow rather than forecasting accuracy alone.

\begin{figure*}[t]
\centering
\includegraphics[width=0.96\textwidth]{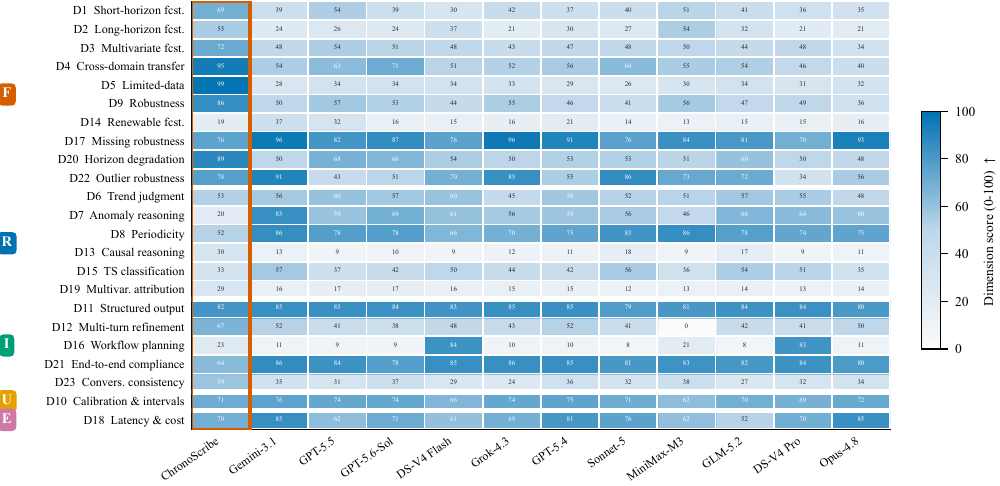}
\caption{Normalized scores across 23 dimensions.}
\label{fig:heatmap}
\end{figure*}

ChronoScribe ranks first at every tested forecasting weight from $0.40$ to $0.80$. \mbox{Figure~\ref{fig:heatmap}} shows normalized scores across 23 dimensions; darker cells indicate higher scores. Its margin over the runner-up grows from $4.9$ points at $w_{\mathrm{F}}=0.40$ to $10.6$ points at the primary weight of $0.60$ and $15.5$ points at $0.80$.

\subsection{Analysis}

\begin{figure}[!t]
\centering
\includegraphics[width=\columnwidth]{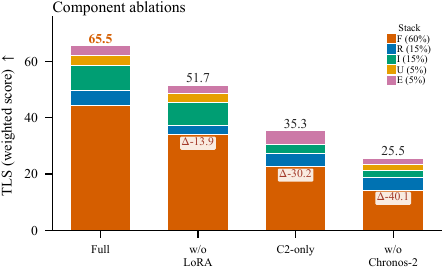}
\caption{TLS decomposition under component ablations.}
\label{fig:ablation}
\end{figure}

TLS falls from $65.5$ to $51.7$ without LoRA adaptation and to $25.5$ without Chronos-2. \mbox{Figure~\ref{fig:ablation}} shows the component ablations. The LoRA ablation changes both the directive and response interfaces, so its effect cannot be attributed to context selection alone. Removing Chronos-2 requires the language model to forecast directly; the resulting drop shows that direct language-model forecasting is not an adequate substitute for the numerical specialist in this setting. C2-only reaches a TLS of $35.3$. Because it omits language-side routing and communication, we treat it as an incomplete-workflow control rather than an estimate of the controller's additive contribution.

The controller returns a parseable directive with the protocol-defined route for $95.9\%$ of requests. On forecasting requests, its horizon field matches the supplied horizon in every case. Because the horizon is fixed by the task, this result measures constraint compliance rather than autonomous horizon inference.

We also audit 100 randomly sampled specialist-path responses using a rule-based verifier. Across 230 extracted claims, agreement with deterministic forecast evidence is $87.4\%$, and 89 responses pass every applicable check. Agreement ranges from $79.6\%$ for numerical values under a $15\%$ relative-error tolerance to $92.1\%$ for directional claims. This audit measures evidence traceability, not forecast accuracy or calibration; the supplementary material provides the complete verification rules.

\section{Discussion and Limitations}

ChronoScribe makes numerical provenance auditable, but the specialist path adds orchestration and an extra LLM call; latency depends on horizon, response length, and backend. We evaluate one LLM--forecaster pairing, Simplified Chinese prompts, and benchmark instruction families, limiting generalization. Although adaptation and evaluation data are disjoint, Chronos-2 was pretrained on public and synthetic corpora, so exposure to individual evaluation series cannot be ruled out~\cite{ansari2025chronos2}. Evidence conditioning improves traceability but cannot guarantee grounded decoding. Lower numerical-claim agreement motivates constrained rendering or verifier-based revision for high-stakes use.

\section{Conclusion}

ChronoScribe applies semantic--numerical specialization to instruction-driven time series forecasting. An adapted LLM interprets requests and proposes structured directives, which deterministic rules convert to executable settings. The frozen TSFM alone produces numerical trajectories; deterministic summaries ground the final response. Across 23 dimensions and 30 public datasets, ChronoScribe achieves the lowest error on all five primary point-error metrics, including fixed-context controls, while the best direct LLM remains $2.0$--$2.7\times$ worse on each. Its TLS of $65.5$ exceeds the strongest of eleven proprietary direct-LLM baselines by $10.6$ points. Routing and grounding audits indicate that the directive and response interfaces perform reliably in most cases. The design retains a single source for forecast values while using the LLM for interpretation, configuration, and communication.

\nocite{makridakis2022m5,lavin2015nab,hundman2018spacecraft,su2019omnianomaly,abdulaal2021psm,mathur2016swat,laptev2015yahoos5,bagnall2018uea,dau2019ucr,dempster2025monster}
\clearpage
\bibliography{ChronoScribe}

\end{document}